\documentclass[a4paper, 11pt]{amia}
\usepackage{graphicx}
\usepackage[labelfont=bf]{caption}
\usepackage[superscript,nomove]{cite}
\usepackage{color}
\usepackage{booktabs}
\usepackage{hyperref}
\usepackage{tabularx,ragged2e,booktabs,caption}
\usepackage[toc,page]{appendix}
\usepackage{mwe}
\usepackage{subcaption}
\usepackage{amsmath}
\usepackage{commath}
\usepackage{listings}
\usepackage{tablefootnote}
\usepackage[utf8]{inputenc}

\newcolumntype{C}[1]{>{\Centering}m{#1}}

\newcolumntype{L}[1]{>{\raggedleft}m{#1}}

\newcolumntype{R}[1]{>{\raggedright}m{#1}}

\begin{document}

\title{Building Deep Learning Models to Predict Mortality in ICU Patients}
\author{Huachuan Wang, Yuanfei Bi}

\institutes{
    George Washington University, Washington, DC, USA\\
}

\maketitle

\section{Abstract}
Mortality prediction in intensive care units (ICUs) is considered one of the critical steps for efficiently treating patients in serious condition. As a result, various prediction models have been developed to address this problem based on modern electronic healthcare records (EHR). However, it becomes increasingly challenging to model such tasks as time-series variables because some laboratory test results such as heart rate and blood pressure are sampled with inconsistent time frequencies. In this paper, we propose several deep learning models using the same features as the SAPS-II score\cite{le1993new}. To derive insight into the proposed models' performance, several experiments have been conducted based on the well-known clinical dataset Medical Information Mart for Intensive Care III (MIMIC-III, v1.4)\cite{mimiciii}. The prediction results demonstrate the proposed models' capability in terms of precision, recall, F1-score, and area under the receiver operating characteristic curve (AUC).

\section{Introduction}
Mortality prediction in Intensive Care Units (ICUs) wards in the hospital where specially trained physicians provide support to the most severely ill patients. However, it is a common challenge that physicians do not have intelligent tools to process a massive amount of modern electronic healthcare records (EHR). The accurate and reliable mortality prediction for ICU patients is crucial for physicians to assess the severity of illness, determine appropriate levels of care, and provide radical life-saving treatment. Patients are monitored closely within ICUs to ensure any deterioration is detected and corrected before it becomes fatal. As a result, there is an increasingly large amount of ICUs data in EHR. Today, deep learning models (aka Deep Neural Networks) have revolutionized many fields such as natural language processing (NLP), voice recognition, and computer vision, and are increasingly adopted in clinical healthcare fields.

This paper aims to develop a deep learning model that can identify patients hospitalized in the ICUs at high risk for death during the ICU stay based on the EMR dataset accumulated by the first 48 hours of the first ICU admission. We propose a method to extract both sequential and non-sequential features from the MIMIC-III (v1.4) database\cite{mimiciii} and build several recurrent neural network (RNN) models to predict hospital mortality, i.e., death inside the hospital. 

The rest of the paper is organized as follows:
In Section \ref{sec:review}, we present a literature review on the related studies.
In Section \ref{sec:data}, we provide a basic statistics of MIMIC-III (v1.4) dataset.
In Section \ref{sec:method}, we describe the pre-processing step we employed to obtain the features and the proposed RNN models.
The experimental results are presented and discussed in Section \textit{Results} and \textit{Discussion}, respectively. We conclude with summary in Section \textit{Conclusion}.

\section{Related Work}\label{sec:review}

Recent advances and success of machine learning and deep learning have facilitated the adoption of these models into ICU patients' mortality prediction tasks. Early work\cite{doig1993modeling, hanson2001artificial, silva2006mortality} showed that machine learning models obtain good results on mortality prediction in ICUs. Recently, an ensemble technique called Super Learner (SL) is proposed to offer improved performance of mortality prediction in ICU patients\cite{pirracchio2016mortality}. Among a given set of candidate algorithms, the SL technique builds an aggregate algorithm as the candidate algorithms' optimally weighted combination. Their work has demonstrated that machine learning models outperform the prognostic scores.

With freely-available datasets such as MIMIC-III, the development of novel models for mortality prediction is gaining increased attention. Lee et al.\cite{lee2015personalized} demonstrated a personalized 30-day mortality prediction model by analyzing similar past patients. Johnson et al.\cite{johnson2017reproducibility} compared multiple published mortality prediction works against gradient boosting and logistic regression model using a simple set of features extracted from MIMIC-III dataset. Recently, researchers have attempted to applied deep learning-based methods to EHR to utilize its ability to learn complex patterns from data. Dabek et al. showed that a neural network model could improve the prediction of several psychological conditions such as anxiety, depression, and behavioral disorders\cite{dabek2015neural}. Che et al.\cite{che2018recurrent} developed a novel recurrent neural network (RNN) model based on Gated Recurrent Unit (GRU), which demonstrates promising performance for ICU mortality prediction. Some RNN models with LSTM units are also proposed and compared with baseline models to show better ICU mortality prediction accuracy\cite{choi2016doctor, ge2018interpretable, zhu2018predicting, zheng2018using}.

\section{Data}\label{sec:data}
MIMIC-III (v1.4)\cite{mimiciii} is a publicly available critical care database maintained by the Massachusetts Institute of Technology (MIT). This database integrates clinical data of over 40,000 patients admitted to ICUs of the Beth Israel Deaconess Medical Center from 2001 to 2012. MIMIC-III consists of 26 relational tables, where 16 of them contain timestamped event information. \autoref{tab:stat} shows the statistics of MIMIC-III (v1.4) dataset. In this project, we will focus on the ICU-related data of adult patients.

\vspace*{16pt}
\begin{minipage}{\linewidth}
\begin{center}
\captionof{table}{Summary statistics of MIMIC-III (v1.4) dataset.}\label{tab:stat} 
\begin{tabular}{R{3.6in}  L{1.4in}}\toprule[1.5pt]
  \# of patients & 46520 \tabularnewline
  \# of adult patients \footnote{Adults: $\geq$ 16 years old.} & 38597 \tabularnewline
  Median age of adult patients & 65.8 years \tabularnewline
  In-hospital mortality of adult patients & 11.5\% \tabularnewline
  \# of admissions & 58976 \tabularnewline
  \# of ICU stays & 61532 \tabularnewline
  \# of ICU stays of adult patients & 53423 \tabularnewline
  \# of long ICU stays \footnote{Long ICU stays: $\geq$ 4 hours.} of adult patients & 53133 \tabularnewline
  \# of the first long ICU stay of adult patients & 38418 \tabularnewline
  Avg. length of long ICU stays of adult patients & 4.17 days  \tabularnewline
  Avg. length of ICU stays of adult patients & 4.14 days  \tabularnewline
  Avg. length of the first long ICU stays of adult patients & 4.07 days  \tabularnewline
\bottomrule[1.25pt]
\end {tabular}
\end{center}
\par
\smallskip
\end{minipage}

\section{Methodology}\label{sec:method}
\subsection{Problem Definition}\label{subsec:prob}
\begin{figure} [hbt!]
\centering
\begin{tabular}{cccc}
\includegraphics[width=0.8\textwidth]{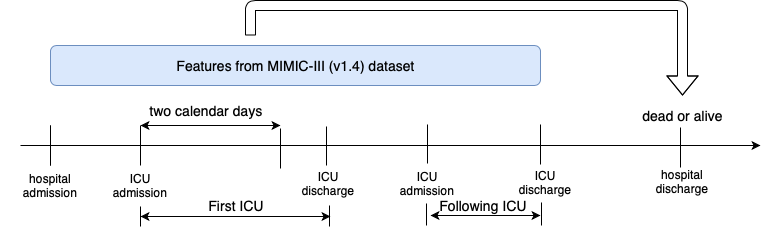}\\
\end{tabular}
\caption{ICU mortality prediction problem}
\label{fig:figure}
\end{figure}

The model we proposed to identify patients hospitalized in the ICU is based on the EMR data accumulated by the first 48 hours into the first ICU stay, as illustrated in \autoref{fig:figure}. Here for each patient, we exclude readmissions of ICU stays, which can prevent possible information leakage in subsequent analysis. Moreover, we choose the prediction time point as the first 48 hours into the first ICU stay because empirical assessment shows that it is impossible to predict ICU mortality accurately without enough data accumulated. 

\vspace{4mm}

\subsection{Cohort Selection}
We use two sets of inclusion criteria to select the ICU stays. First, as mentioned in Section \ref{subsec:prob}, we exclude readmissions of ICU stays. Second, we choose ICU stays that meet the following criteria: age of patient $\geq$ 16 years at the time of ICU admission, and the ICU stay is longer than 48 hours.

\subsection{Data Cleaning}
Due to noise, missing values, outliers, or incorrect records, the data extracted from the MIMIC-III database has lots of erroneous entries. Therefore we need to identify and handle these inconsistent or erroneous records. First, we observed that there is inconsistency in the measure units of some variables. For example, the body temperature is measured in either Fahrenheit or Celsius units. Second, some numerical values are missing or recorded as error texts. Third, some variables have multiple values recorded at the same time. We addressed these issues by following procedures.
\begin{itemize}
  \item To handle inconsistent units in body temperature records, we represent all data in Fahrenheit unit.
 \item For missing records, there are two circumstances. First, if the record is only missing occasionally with 48 hours, we do forward imputation and backward imputation. Second, if there is no such data for a period of 48 hours, we take the average value of that variable of all patients.
 \item For multiple records of the same variable in an hour, we randomly pick one value as minimal changes.
\end{itemize}

\subsection{Feature Selection and Extraction}
We extract data from following tables: \textit{admissions}, \textit{services}, \textit{outputevents}, \textit{chartevents}, \textit{icustays}, \textit{labevents} and \textit{diagnoses\_icd}, etc, because they provide the most relevant clinical features of ICU stays. To enable an exhaustive feature map that can measure the severity of disease for patients admitted to ICUs efficiently, we select the same set of features that are used in the calculation of the SAPS-II score, which consists of the 17 features. \autoref{tab:feature} lists all the 17 processed features and their corresponding entries in the MIMIC-III database tables.

\vspace{2mm}
\begin{minipage}{\linewidth}
\centering
\captionof{table}{17 features used in SAPS-II scoring system}\label{tab:feature}
\begin{tabular}{ R{1.8in} R{0.8in} R{2.4in} L{0.7in}}\toprule[1.5pt]
\bf Features & \bf ItemID & \bf Item Name & \bf Table \tabularnewline \midrule
  glasgow coma scale & 723 & GCSVerbal  & chartevents \tabularnewline
  & 454 & GCSMotor & chartevents\tabularnewline
  & 184  & GCSEyes & chartevents \tabularnewline
  & 223900  & Verbal Response & chartevents \tabularnewline
  & 223901  & Motor Response & chartevents \tabularnewline
  & 220739  & Eye Opening & chartevents \tabularnewline \midrule
  systolic blood pressure & 51  & Arterial BP [Systolic] & chartevents \tabularnewline
  & 442  & Manual BP [Systolic] & chartevents \tabularnewline
  & 455  & NBP [Systolic] & chartevents \tabularnewline
  & 6701 & Arterial BP \# 2 [Systolic] & chartevents \tabularnewline
  & 220179 & Non Invasive Blood Pressure systolic & chartevents \tabularnewline
  & 220050  & Arterial Blood Pressure systolic & chartevents \tabularnewline  \midrule
  heart reate & 211  & Heart Rate & chartevents \tabularnewline
  & 220045  & Heart Rate & chartevents \tabularnewline  \midrule
  body temperature & 678  & Temperature F & chartevents \tabularnewline
  & 223761  & Temperature Fahrenheit & chartevents \tabularnewline
  & 676  & Temperature C & chartevents \tabularnewline 
  & 223762  & Temperature Celsius & chartevents \tabularnewline \midrule
  pao2 / fio2 & 50821  & PO2 & labevents \tabularnewline
  & 50816  & Oxygen &  labevents \tabularnewline 
  & 223835    & Inspired O2 Fraction (FiO2) &  chartevents \tabularnewline
  & 3420    & FiO2 & chartevents \tabularnewline
  & 3422    & FiO2 (Meas) & chartevents \tabularnewline
  & 190    & FiO2 set & chartevents \tabularnewline \midrule
  
  urine output & 40055  & Urine Out Foley & outputevents \tabularnewline
  &43175    & Urine & outputevents \tabularnewline
  &40069    & Urine Out Void & outputevents \tabularnewline
  &40094    & Urine Out Condom Cath & outputevents \tabularnewline
  &40715    & Urine Out Suprapubic & outputevents \tabularnewline
  &40473    & Urine Out IleoConduit & outputevents\tabularnewline
  &40085    & Urine Out Incontinent & outputevents\tabularnewline
  &40057    & Urine Out Rt Neophrostomy & outputevents \tabularnewline
  &40056    & Urine Out Lt Neophrostomy & outputevents\tabularnewline
  &40405    & Urine Out Other & outputevents\tabularnewline
  &40428    & Orine Out Straight Cath & outputevents\tabularnewline
  &40086    & Urine Out Ureteral Incontinent  & outputevents\tabularnewline
  &40096    & Urine Out Ureteral Stent \# 1 & outputevents\tabularnewline
  &40651    & Urine Out Ureteral Stent \# 2 & outputevents\tabularnewline
  &226559    & Foley & outputevents\tabularnewline
  &226560    & Void & outputevents\tabularnewline
  &226561    & Condom Cath & outputevents\tabularnewline 

\end {tabular}\par
\smallskip
\end{minipage}

\vspace{2mm}
\begin{minipage}{\linewidth}
\centering
\captionof*{table}{\textbf{Table 2:} 17 features used in SAPS-II scoring system}
\begin{tabular}{ R{1.8in} R{0.8in} R{2.4in} L{0.8in}}\toprule[1.5pt]
\bf Features & \bf ItemID & \bf Item Name & \bf Table \tabularnewline \midrule
  &226584    & Ileoconduit & outputevents \tabularnewline
  &226563    & Suprapubic & outputevents \tabularnewline
  &226564    & R Nephrostomy & outputevents \tabularnewline
  & 226565 & L Neophrostomy  & outputevents \tabularnewline
  & 226567 & Straight Cath & outputevents \tabularnewline
  & 226557  & R Ureteral Stent & outputevents \tabularnewline
  & 226558  & L Ureteral Stent & outputevents \tabularnewline
  & 227488  & GU Irrigant Volume In  & outputevents \tabularnewline
  & 227489  & GU Irrigant/Urine Volume Out & outputevents \tabularnewline \midrule
  serum urea nitrogen level & 51006  & Urea Nitrogen & labevents \tabularnewline \midrule
  
  white blood cells count & 51300  & WBC Count & labevents \tabularnewline
  & 51301  & White Blood Cells & labevents \tabularnewline \midrule
  
  serum bicarbonate level & 50882 & BICARBONATE &  labevents \tabularnewline \midrule
  
  sodium level & 950824 & Sodium White Blood & labevents \tabularnewline
  & 50983  & Sodium & labevents \tabularnewline  \midrule
  
  potassium level & 50822  & Potassium, whole blood & chartevents \tabularnewline
  & 50971  & Potassium & chartevents \tabularnewline  \midrule
  
  bilirubin level & 50885  & Bilirubin Total & labevents \tabularnewline \midrule
  
  age & -  & intime & icustays  \tabularnewline
  & -  & dob  &  patients \tabularnewline \midrule
  
  immunodeficiency syndrome & -  & icd9\_code &  diagnoses\_icd \tabularnewline \midrule
  
  hematologic malignancy & -  & icd9\_code & diagnoses\_icd \tabularnewline \midrule
  metastatic cancer & -  & icd9\_code & diagnoses\_icd \tabularnewline \midrule
  admission type & -    & curr\_service & services \tabularnewline
  &     & ADMISSION\_TYPE & admissions \tabularnewline
  \bottomrule[1.25pt]
\end {tabular}\par
\smallskip
\end{minipage}
\vspace{2mm}

The 17 features in \autoref{tab:feature} can be divided into two categories: non-sequential features such as chronic diseases, admission types and age, and sequential features that represent time-series patient characteristic such as blood pressure, heart rate, and body temperature, etc. For each patient admitted into ICU, each time-series feature is sampled every 1 hour so that a 48$\times$13 matrix represents the time-series information for each patient.

\subsection{Deep Learning Models}
Recently, deep learning models have demonstrated promising performance in mortality prediction of ICU patients. Deep learning models consist of a layered, hierarchical architecture of neurons for learning and representing data. One of the main advantages of the deep learning models is their ability to learn good features from raw data automatically and significantly reduce handcrafted feature engineering. Some recent works have demonstrated that deep learning models achieve state-of-the-art performance in health-related fields, such as ICU mortality prediction\cite{johnson2017reproducibility}, phenotype discovery\cite{che2015deep} and disease prediction\cite{hinton2015distilling}. We applied the RNN model in this work, which is appropriate for modeling sequence and time-series data.  

\subsubsection{Implementation Details}
Here we implemented a basic 3-layer LSTM model in PyTorch\cite{paszke2017automatic}. The model is trained with Adam optimizer with a learning rate of 0.001. The batch size is 32, and the max epoch number is 10. Early stopping with the best weight is applied during training. We randomly sample 20\% of the patients for the test set and 20\% for the validation set. The remaining 60\% of the patients are used during training.

\subsubsection{Evaluation Metrics}
As the ICU mortality is a binary classification problem, we choose \textit{Precision}, \textit{Recall}, \textit{F1} and \textit{AUC} to evaluate our models. 

\section{Prediction Results}
\begin{figure} [hbt!]
\centering
\begin{tabular}{cccc}
\includegraphics[width=0.65\textwidth]{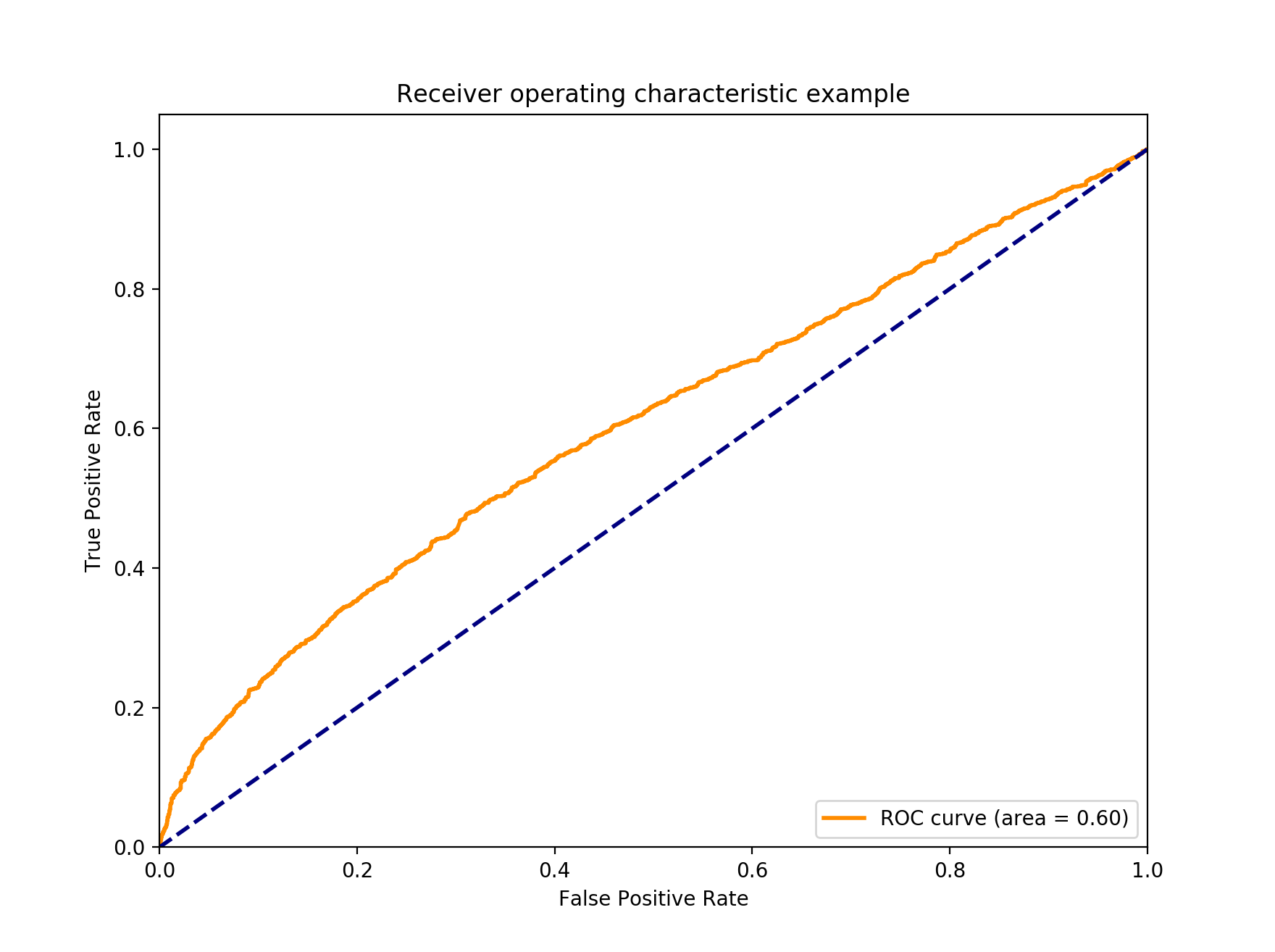} \\
\end{tabular}
\caption{The ROC of the RNN-LSTM model.}
\label{fig:roc}
\end{figure}

In this paper, we compared the RNN-LSTM-based model with a logistic regression model with L2 regularization. The logistic regression model's input feature values are measured at the last hour of the 48 hours window. The metrics results of the basic LSTM model and the comparison logistic regression model are reported in \autoref{tab:result} and the receiver operating characteristic (ROC) curve of the RNN-LSTM model is in \autoref{fig:roc}. From \autoref{tab:result}, RNN-LSTM model consistently outperforms the baseline logistic regression model. On the test dataset, the AUC of the RNN-LSTM model is higher than logistic regression by 4\%. \autoref{fig:roc} shows that a basic LSTM model can achieve good performance in mortality prediction, which implies a promising future of deep learning models in health-related projects.

\vspace*{16pt}
\begin{minipage}{\linewidth}
\centering
\captionof{table}{Metrics evaluation of different models.}\label{tab:result}
\begin{tabular}{ R{1.3in} C{0.6in} C{0.6in} C{0.6in} C{0.6in} C{0.6in}}\toprule[1.5pt]
\bf Model & \bf Precision & \bf Recall & \bf F1 & \bf AUC \tabularnewline \midrule
 RNN-LSTM model & 0.620 & 0.711 & 0.662 & 0.600 \tabularnewline
 Logistic Regression & 0.610 & 0.650 & 0.620 & 0.560 \tabularnewline
\bottomrule[1.25pt]
\end {tabular}\par
\smallskip
\end{minipage}

\section{Discussion}
In this study of nearly 40,000 ICU stays, we found that an RNN-LSTM based model can take advantage of the sequential nature of time-series features to achieve higher accuracy in identifying patients at high risk of death to some common approaches such as logistic regression. This finding demonstrates that it is important to perform a sequential clinical data analysis because an abnormal change in a key physiologic measurement may signal potential clinical deterioration, even if the absolute value is not in a critical zone yet. Our research work sheds new light to empower deep learning in the health-related project.

Although we used the same features as SAPS-II calculation in this work, it is worth mentioning that identifying efficient features to predict ICU survival is not trivial. This, therefore, remains to be an important direction for future research.

This present study also has several other limitations. First, the MIMIC-III dataset is collected from a single intuition, so our findings may not be generalizable to other clinical or geographic settings. The data from a medical ICU may not apply to other ICU categories. Second, some other data available from the MIMIC-III dataset, such as fluid balance and monitor data, have not been incorporated into our model. Future work will focus on aggregating these additional data and quantifying their impact on prediction accuracy. Third, the RNN-LSTM model implemented in this work has only three layers that lack the capability to capture sequential and non-sequential features efficiently.

\section{Conclusion}
To conclude, in this work, we propose to apply deep learning models into mortality prediction of ICU patients on the MIMIC-III (v1.4) dataset. We preprocess data and extract features that have been used in SAPS-II. These features include both sequential and non-sequential data, which better reflects patients' psychological conditions. Then we implement and train a basic RNN-LSTM model and compare its prediction performance with that of a logistic regression model. Our result shows that the basic RNN-LSTM model can stably exceed the accuracy of a "traditional" logistic regression model. Our deep learning model's significance includes 1) by effectively capturing fluctuations in time-series features, it could give clinicians an early sense of the patient's mortality status; and 2) it could be used to help allocate ICU resources more efficiently.

In the future, our work can be extended in several directions. For example, 1) more sophisticated data preprocessing steps and deep learning models will be conducted to capture the characteristics of the massive MIMIC-III datasets, and 2) more extensive ICU datasets will be employed to evaluate and improve our models.

\newpage

\bibliographystyle{unsrt}
\bibliography{main}

\end{document}